\documentclass{article}
\usepackage{iclr2027_conference,times}

\usepackage{amsmath,amsfonts,bm}

\def\eqref#1{equation~\ref{#1}}

\def\1{\bm{1}}

\DeclareMathAlphabet{\mathsfit}{\encodingdefault}{\sfdefault}{m}{sl}
\SetMathAlphabet{\mathsfit}{bold}{\encodingdefault}{\sfdefault}{bx}{n}

\usepackage{amsmath,amssymb}
\usepackage{graphicx}
\usepackage{wrapfig}
\usepackage{float}
\usepackage{booktabs}
\usepackage{multirow}
\usepackage{array}
\usepackage{microtype}
\usepackage{xcolor}
\usepackage{colortbl}
\usepackage{hyperref}
\usepackage{url}

\newcommand{\method}{PMOPD}
\newcommand{\best}[1]{\textbf{#1}}

\title{PMOPD: Task Ordering, Cycling, and Parameter-Update Subspace Protection in Multi-Teacher On-Policy Distillation}

\author{
  \makebox[0.49\linewidth][c]{%
    \begin{tabular}[t]{c}
      \textbf{Youzhi Liu} \\
      \textmd{Ant Group} \\
      \textmd{\texttt{liuyouzhi22@mails.ucas.ac.cn}}
    \end{tabular}%
  }\And
  \makebox[0.49\linewidth][c]{%
    \begin{tabular}[t]{c}
      \textbf{Ruobing Zheng}\thanks{Corresponding authors.} \\
      \textmd{Ant Group} \\
      \textmd{\texttt{zrb915@gmail.com}}
    \end{tabular}%
  }\AND
  \makebox[0.3\linewidth][c]{%
    \begin{tabular}[t]{c}
      \textbf{Boyuan Tong} \\
      \textmd{Ant Group}
    \end{tabular}%
  }\And
  \makebox[0.3\linewidth][c]{%
    \begin{tabular}[t]{c}
      \textbf{Tianqi Li} \\
      \textmd{Ant Group}
    \end{tabular}%
  }\And
  \makebox[0.3\linewidth][c]{%
    \begin{tabular}[t]{c}
      \textbf{Pingqi Li} \\
      \textmd{Ant Group}
    \end{tabular}%
  }\AND
  \makebox[0.3\linewidth][c]{%
    \begin{tabular}[t]{c}
      \textbf{Hanbo Bi} \\
      \textmd{Ant Group}
    \end{tabular}%
  }\And
  \makebox[0.3\linewidth][c]{%
    \begin{tabular}[t]{c}
      \textbf{Yi Yuan} \\
      \textmd{Ant Group}
    \end{tabular}%
  }\And
  \makebox[0.3\linewidth][c]{%
    \begin{tabular}[t]{c}
      \textbf{Jingdong Chen} \\
      \textmd{Ant Group}
    \end{tabular}%
  }
}

\iclrfinalcopy 

\begin{document}
\maketitle
\fancyhead{}
\renewcommand{\headrulewidth}{0pt}

\begin{abstract}
Multi-teacher on-policy distillation (MOPD) has emerged as a popular post-training paradigm for integrating specialized capabilities in frontier language models. Existing OPD research has primarily focused on optimizing single-task distillation through objective design, distillation scope, and teacher signal construction, whereas MOPD must aggregate multiple capabilities in shared parameters and address the resulting capability seesaw, in which improving one domain suppresses capabilities acquired from another. Inspired by the distinctive update geometry of OPD, we find that parameter updates from different tasks rapidly concentrate in their respective low-dimensional subspaces during MOPD, providing a direct geometric basis for identifying and controlling cross-task interference. We therefore propose \method{} (Projection-based Multi-Teacher On-Policy Distillation), which constructs subspace memories from the cumulative parameter displacements of different tasks and projects both gradients and optimizer updates to remove components that interfere with protected task directions. We further develop a lightweight conflict probe to characterize task interactions and guide task ordering, together with a cycling strategy that balances subspace estimation and timely task revisitation. Experiments on representative Code, Reason, and Math tasks show that \method{} improves every evaluated capability over MOPD, raising the average score across the three tasks by 2.54 points on Qwen2.5-7B and 2.09 points on Llama-3.1-8B. These consistent gains establish geometry-aware optimization as an effective and transferable approach to balanced multi-teacher distillation.

\end{abstract}

\section{Introduction}
\label{sec:intro}
Multi-teacher on-policy distillation (MOPD) is increasingly used as a post-training paradigm for integrating specialized capabilities into a single language model \citep{ma2026mopd,deepseek2026v4,mimo2026v2flash,kimi2026k3}. Starting from a shared base model, separate teachers can be optimized for domains such as mathematics, reasoning, and code, after which their behaviors are distilled into one deployable student. Compared with maintaining an ensemble of specialists, MOPD avoids inference-time routing and multi-model serving costs while retaining on-policy supervision: teachers provide token-level targets on trajectories sampled from the student itself \citep{agarwal2024gkd,ma2026mopd}.

Most existing OPD studies, however, optimize distillation in single-task settings through objective design, distillation scope, and teacher signal construction \citep{li2026rethinkingopd,yu2026openmopd}. MOPD introduces a distinct optimization challenge: multiple teachers act on the same student parameters, and their objectives need not be compatible. Consequently, improving one capability can suppress another, producing a capability seesaw \citep{ma2026mopd,yu2026openmopd}. Standard task mixing reduces the interval between domains but neither identifies conflicting update directions nor protects useful changes already induced by earlier teachers. MOPD therefore requires the student to balance stability and plasticity by preserving acquired capabilities while retaining enough freedom to absorb new ones.

We approach this problem through the update geometry of OPD. Instead of treating each noisy mini-batch gradient independently, we examine the cumulative parameter displacement over a task block. Its dominant singular subspace stabilizes early, consistent with recent analyses of OPD update geometry \citep{shen2026geometry}. It also exhibits high consistency across independent shards of the same task and low overlap across tasks. This suggests that a compact subspace can summarize the main directions used by a teacher and that the overlapping components of later updates provide a tractable target for interference control without globally freezing the parameter space.

Based on this observation, we propose \method{} (Projection-based Multi-Teacher On-Policy Distillation), which constructs per-matrix subspace memories from the cumulative parameter displacements of completed task blocks. When learning a subsequent task, it projects both the gradient and the preconditioned optimizer update to remove components that interfere with protected task directions. The second projection is necessary because the element-wise adaptive preconditioning used by optimizers such as Adafactor \citep{shazeer2018adafactor} can rotate an already projected gradient back toward the protected subspaces. Memories are rebuilt in each cycle, allowing the protected geometry to track the current trajectory without accumulating an unbounded archive.

\begin{figure*}[t]
\centering
\includegraphics[width=0.96\textwidth]{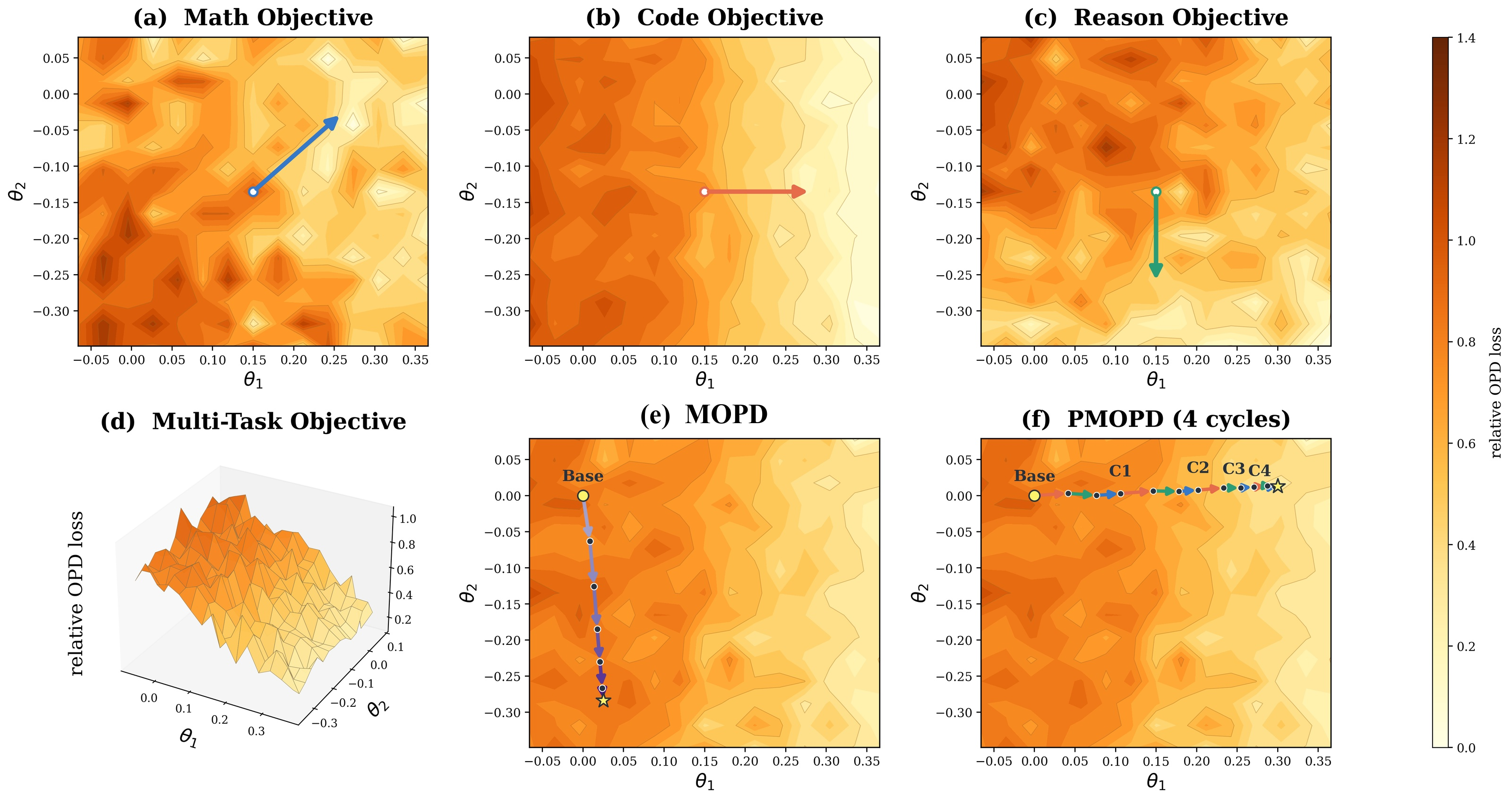}
\caption{Visualization of task-preferred update directions and the optimization trajectories of MOPD and \method{}. The axes $\theta_1$ and $\theta_2$ define a rank-2 PCA plane derived from actual checkpoint displacements. (a-c) show the normalized Math, Code, and Reason OPD losses. The arrows indicate their preferred next-update directions from a shared reference checkpoint. (d) shows the equally weighted multi-task relative OPD loss, while (e-f) overlay the actual MOPD and four-cycle \method{} trajectories on this joint landscape. Lighter colors indicate lower relative loss and darker colors indicate higher relative loss. The task-specific arrows differ substantially at the same checkpoint, and MOPD does not consistently descend on the joint landscape. By correcting updates against protected task directions, \method{} follows a more stable trajectory toward a shared low-loss region.}
\label{fig:teaser}
\vspace{-1.5\baselineskip}
\end{figure*}

Figure~\ref{fig:teaser} provides a geometric view of this interference. At the same reference checkpoint, Math, Code, and Reason favor markedly different update directions, so an update that benefits one objective need not reduce the others. Correspondingly, the measured MOPD trajectory fluctuates on the joint loss landscape, whereas \method{} more consistently enters a shared low-loss region by
correcting updates against previously identified task-update subspaces.

MOPD also depends on temporal organization. Because pairwise interference is directional, we measure both directions and aggregate them with a lightweight conflict probe to rank tasks before training. We cycle through teachers to balance timely revisitation against reliable block-displacement estimates. In our setting, the probe selects Code$\rightarrow$Reason$\rightarrow$Math, and four cycles perform best.

Across Qwen2.5-7B and Llama-3.1-8B, \method{} raises the average score across the three tasks over MOPD by 2.54 and 2.09 points, respectively, while improving every evaluated task in both model families. This uniform advantage shows that \method{} strengthens joint capability integration instead of redistributing performance among domains.

\paragraph{Contributions.} Our contributions are threefold:
\begin{itemize}
    \item We characterize MOPD interference geometrically, showing that cumulative OPD updates form early-stabilizing, low-dimensional subspaces that are consistent within tasks and exhibit low cross-task similarity.
    \item We introduce \method{}, which constructs subspace memories from realized task displacements and protects them by projecting both gradients and adaptive-optimizer updates.
    \item We develop a lightweight conflict probe and cycling strategy for organizing task updates, and demonstrate consistent improvements across two model families and three capability domains.
\end{itemize}

\vspace{-0.45\baselineskip}
\section{Related Work}
\label{sec:related}
\vspace{-0.35\baselineskip}
\subsection{Knowledge distillation and on-policy supervision}
Knowledge distillation transfers teacher behavior through distribution matching, generated sequences, or policy supervision \citep{hinton2015distilling,kim2016sequence,rusu2015policy}. OPD instead queries teachers on student trajectories and supplies token-level targets for visited prefixes, reducing the mismatch between training and inference states \citep{ross2011reduction,agarwal2024gkd}. Recent systems use multiple teachers to integrate domain specialists \citep{ma2026mopd,deepseek2026v4,mimo2026v2flash,kimi2026k3}, while concurrent work studies OPD objectives and practical interference \citep{li2026rethinkingopd,yu2026openmopd}. We study this interference through shared-parameter update geometry.

\vspace{-0.35\baselineskip}
\subsection{Multi-task optimization and capability integration}
Conflicting multi-task gradients motivate projection, common-descent, and magnitude-balancing methods \citep{yu2020pcgrad,sener2018multi,chen2018gradnorm}. PCGrad removes pairwise conflicting components, multi-objective optimization seeks common descent directions, and GradNorm balances task scales. These methods operate on instantaneous gradients. In contrast, \method{} stores realized OPD block displacements, including clipping and adaptive-optimization effects, to constrain later updates.

Prediction-time ensembles retain specialist models \citep{breiman1996bagging,lakshminarayanan2017deep,huang2017snapshot}, whereas model soups and task arithmetic combine independently trained parameters \citep{wortsman2022soups,ilharco2023task}. Such parameter-space combinations can conceal conflicts until merging and do not expose the combined model to its generated states. \method{} instead integrates capabilities under on-policy supervision and directly constrains the optimization path.

\vspace{-0.35\baselineskip}
\subsection{Continual learning and low-dimensional update geometry}
Continual-learning methods preserve prior tasks through parameter penalties, episodic constraints, or activation-derived subspace projection \citep{kirkpatrick2017overcoming,lopezpaz2017gem,chaudhry2019agem,saha2021gpm}. GPM extracts protected bases from activations. In contrast, \method{} derives per-matrix bases from realized OPD block displacements, applies them to gradients and optimizer updates, and rebuilds memory each cycle.

Neural-network adaptation often lies in low-dimensional spaces, as exploited by intrinsic-dimension methods and LoRA \citep{aghajanyan2021intrinsic,hu2022lora}. Cumulative OPD updates similarly exhibit early subspace locking \citep{shen2026geometry}. We further show strong same-task alignment and low cross-task overlap, making the overlapping directions a tractable target for interference control.

\section{Problem Formulation and Geometric Motivation}
\label{sec:motivation}
\subsection{Multi-teacher on-policy distillation}
Let $\mathcal{T}=\{\text{Math},\text{Reason},\text{Code}\}$ denote the task set, $D_t$ the prompt distribution for task $t$, $\pi_\theta$ the shared student, and $\pi_T^t$ the corresponding specialist teacher. For $x\sim D_t$, the student samples a response $y\sim\pi_\theta(\cdot\mid x)$ and the teacher is evaluated on the same prefixes, following the standard on-policy distillation setup \citep{agarwal2024gkd,ma2026mopd}. A multi-task run minimizes
\begin{equation}
\min_\theta\;\mathbb{E}_{t\sim q,\,x\sim D_t,\,y\sim\pi_\theta}
\left[\mathcal{L}_{\mathrm{OPD}}^t(\theta;x,y)\right],
\label{eq:multi-objective}
\end{equation}
where $q$ is induced by the task schedule. 


\subsection{Task-update subspaces}
For a two-dimensional weight matrix $W$, define the accumulated block update and its singular value decomposition as
\begin{equation}
\Delta W^t=W^{\mathrm{after}}-W^{\mathrm{before}},\qquad
\Delta W^t=U^t\Sigma^t(V^t)^\top.
\label{eq:svd}
\end{equation}
We retain the top-$K$ right-singular vectors $V_{t,K}$. For two blocks $a$ and $b$, we quantify subspace overlap by
\begin{equation}
S_K(a,b)=\frac{1}{K}\left\|V_K(a)^\top V_K(b)\right\|_F^2.
\label{eq:overlap}
\end{equation}
Following prior analysis of OPD update geometry \citep{shen2026geometry}, we set $K=16$ to retain the dominant update directions in a compact basis with low memory and projection costs. With this setting, the subspaces extracted from cumulative displacements after 20\% of training already attain an average similarity of 0.62 to their corresponding final subspaces across Math, Reason, and Code. This result shows that the dominant update subspace for each task emerges early and remains stable throughout OPD, enabling it to serve as a compact geometric representation of task-specific parameter updates.

\begin{wraptable}{r}{0.46\textwidth}
\centering
\vspace{-2.0\baselineskip}
\caption{Similarities among the top-16 OPD update subspaces induced by different training domains. The low cross-domain similarities support selective subspace protection.}
\label{tab:cross-task-subspace}
\vspace{0.15\baselineskip}
\small
\setlength{\tabcolsep}{4.2pt}
\renewcommand{\arraystretch}{1.08}
\begin{tabular}{lccc}
\toprule
 & Math & Code & Reason \\
\midrule
Math   & \cellcolor{blue!62}1.000 & \cellcolor{blue!22}0.151 & \cellcolor{blue!16}0.142 \\
Code   & \cellcolor{blue!22}0.151 & \cellcolor{blue!62}1.000 & \cellcolor{blue!10}0.133 \\
Reason & \cellcolor{blue!16}0.142 & \cellcolor{blue!10}0.133 & \cellcolor{blue!62}1.000 \\
\bottomrule
\end{tabular}
\vspace{-1.0\baselineskip}
\end{wraptable}
We next compare the final task-update subspaces across domains. As shown in Table~\ref{tab:cross-task-subspace}, the similarities among the top-16 update subspaces for Math, Code, and Reason range from $0.133$ to $0.151$, demonstrating that their parameter updates occupy substantially different subspaces. This strong cross-task separation provides the geometric basis for compactly protecting task-specific update directions, while the overlapping components identify the directions in which later task updates interfere with protected capabilities.

\section{Method}
\label{sec:method}

\subsection{Overview}
\begin{figure}[t]
\centering
\includegraphics[width=0.96\textwidth]{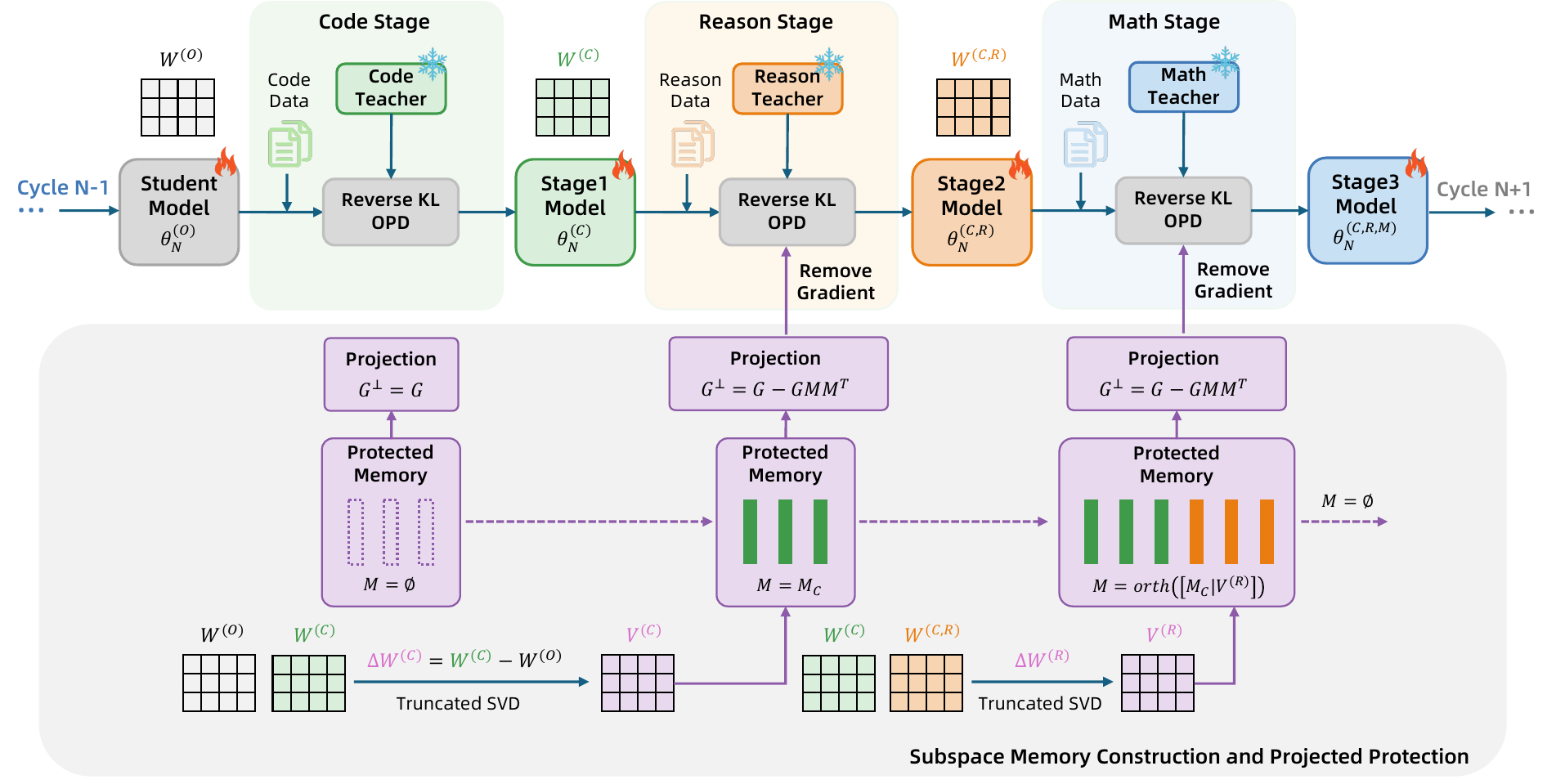}
\caption{Overview of \method{} in cycle $N$. The upper row shows sequential OPD across the Code, Reason, and Math stages, and the lower row shows subspace-memory construction and projected protection. Code is trained with an empty memory. Its block displacement constructs the first protected basis, which is expanded after Reason and used to constrain the Math stage. The final student initializes cycle $N{+}1$, where the memory is rebuilt from new task blocks.}
\label{fig:method-overview}
\vspace{-1.0\baselineskip}
\end{figure}

Our method, \method{}, performs multi-teacher OPD in ordered task blocks. Within each cycle, the first task is learned without a protection constraint and its realized parameter displacement is compressed into compact per-matrix memories. Later tasks are trained after removing the components of both their gradients and optimizer updates that lie in the accumulated memory. The memory is rebuilt from an empty state in every cycle, so it represents the update geometry induced by the current data shards rather than an ever-growing archive across cycles. Figure~\ref{fig:method-overview} illustrates the Code$\rightarrow$Reason$\rightarrow$Math progression within cycle $N$. The upper row follows the shared student through the three task stages, while the lower row shows how completed block displacements construct and expand the subspace memory used to project subsequent gradients. The core procedure contains three components: reverse-KL on-policy distillation, task-update subspace extraction, and dual projection of gradients and Adafactor updates. We select the task order and cycle count with lightweight diagnostics analyzed in Section~\ref{sec:analysis}.

\subsection{Reverse-KL on-policy distillation}
For a prompt $x$ from task $t$, the student first samples a response $y\sim\pi_\theta(\cdot\mid x)$. The task teacher is then evaluated on the same student-generated prefixes $h_i=(x,y_{<i})$. The main experiments minimize reverse KL,
\begin{equation}
\mathcal{L}_{\mathrm{RKL}}^{t}
=\frac{1}{|y|}\sum_{i=1}^{|y|}
\operatorname{KL}\!\left(
\pi_\theta(\cdot\mid h_i)\,\|\,\pi_T^t(\cdot\mid h_i)
\right),
\label{eq:rkl}
\end{equation}
where
\begin{equation}
\operatorname{KL}(\pi_\theta\|\pi_T^t)
=\sum_{v\in\mathcal{V}}\pi_\theta(v\mid h_i)
\log\frac{\pi_\theta(v\mid h_i)}{\pi_T^t(v\mid h_i)}.
\end{equation}
Thus, every teacher supervises states actually visited by the shared student, while the task identity determines which specialized teacher supplies the target distribution.

\subsection{Extracting task-update subspaces}
\method{} performs subspace extraction and projection independently for every two-dimensional trainable parameter matrix, rather than constructing a single unified subspace for the entire model. To keep the notation concise, we describe the operation for one arbitrary matrix and omit its matrix index throughout this section.

Consider a trainable parameter $W\in\mathbb{R}^{m\times n}$. At the beginning and end of a task block, we record $W^{\mathrm{before}}$ and $W^{\mathrm{after}}$, and form the realized displacement
\begin{equation}
\Delta W^t=W^{\mathrm{after}}-W^{\mathrm{before}}.
\label{eq:block-delta}
\end{equation}
Unlike an individual mini-batch gradient, $\Delta W^t$ includes the cumulative effect of the task loss, gradient clipping, and the optimizer over the entire block. We approximate it with a randomized truncated SVD,
\begin{equation}
\Delta W^t\approx U_K^t\Sigma_K^t(V_K^t)^\top,
\qquad K=16.
\label{eq:truncated-svd}
\end{equation}
The columns of $V_K^t\in\mathbb{R}^{n\times K}$ are the dominant input-side directions of the task-induced parameter change: $\Delta W^t v_i=\sigma_i u_i$. We use right rather than left singular vectors because the protection operation acts on the right side of the gradient and optimizer-update matrices.

\subsection{Orthogonal task-subspace memory}
Let $M$ denote the subspace memory associated with the current parameter matrix. After a remembered task block, its new directions are concatenated with the existing basis and orthogonalized by a reduced QR decomposition,
\begin{equation}
A=[M\mid V_K^t]=QR,
\qquad
M\leftarrow Q[:,\,|\operatorname{diag}(R)|>\epsilon],
\label{eq:memory-qr}
\end{equation}
where $\epsilon$ is a small redundancy threshold. QR removes numerically redundant directions and ensures $M^\top M=I$. Consequently, $P=MM^\top$ is the orthogonal projector onto the protected input-side subspace.

\subsection{Gradient and optimizer-update projection}
For an unprojected gradient $G\in\mathbb{R}^{m\times n}$, we decompose
\begin{equation}
G_{\parallel}=GMM^\top,
\qquad
G_{\perp}=G-GMM^\top.
\label{eq:gradient-projection}
\end{equation}
The backward gradient is replaced by $G_{\perp}$ before global gradient-norm clipping. Since $M^\top M=I$, the retained gradient satisfies
\begin{equation}
G_{\perp}M=(G-GMM^\top)M=0.
\end{equation}
Therefore, at the level of this linear map, the projected update does not directly change the transformation along inputs in $\operatorname{span}(M)$.

For a parameter matrix, we measure the fraction of its gradient aligned with the protected memory as
\begin{equation}
r_g=\frac{\|GMM^\top\|_F}{\|G\|_F}.
\label{eq:gradient-removed}
\end{equation}
The global diagnostic reported in our experiments accumulates the squared numerator and denominator norms over all projected parameter matrices before taking their ratio. 

Because Adafactor's adaptive preconditioning can change the direction of an already projected gradient, we additionally project the preconditioned optimizer update:
\begin{equation}
U_{\perp}=U-UMM^\top,
\qquad
W\leftarrow W-U_{\perp},
\label{eq:update-projection}
\end{equation}
where $U$ is the raw Adafactor update after preconditioning.

\section{Experimental Setup}
\label{sec:setup}
\subsection{Models, tasks, and teachers}
We evaluate two independently developed base-model families: Qwen2.5-7B \citep{qwen2025qwen25} and Llama-3.1-8B \citep{dubey2024llama3}. For each family, the Math, Reason, and Code teachers begin from the same base checkpoint and are specialized for their respective domains. Specifically, each teacher is obtained by reinforcement learning (RL) on its domain data, with the training-set sizes reported in the RL column of Table~\ref{tab:data}. This shared initialization makes the integration setting controlled: teacher differences arise from domain-specific RL rather than unrelated pretraining histories.

\begin{wraptable}{r}{0.40\textwidth}
\vspace{-2.0\baselineskip}
\caption{Data configuration for the main experiments.}
\label{tab:data}
\centering
\small
\setlength{\tabcolsep}{5pt}
\begin{tabular}{@{}lrrr@{}}
\toprule
Task & RL & OPD & Test\\
\midrule
Math & 2700 & 600 & 450\\
Reason & 4870 & 600 & 1221\\
Code & 2400 & 600 & 300\\
\bottomrule
\end{tabular}
\vspace{-0.7\baselineskip}
\end{wraptable}
The primary Qwen configuration is summarized in Table~\ref{tab:data}. Math uses the \textnormal{orca\_math} split of NuminaMath-CoT \citep{li2024numinamath}, Reason uses CommonsenseQA \citep{talmor2019commonsenseqa}, and Code uses APPS \citep{hendrycks2021apps}. Each task contributes 600 OPD prompts. Evaluation uses the test sets and counts listed in Table~\ref{tab:data}. Under the selected four-cycle schedule, each task block contains 150 prompts.

\subsection{Baselines and evaluation}
We compare against: (i) the shared base model; (ii) each single-domain teacher; (iii) parameter merging, which averages compatible specialist parameters; (iv) MOPD, the reported multi-teacher OPD baseline that mixes examples from different tasks within each mini-batch \citep{ma2026mopd}; (v) BB-MOPD, which performs between-batch mixing by interleaving task-homogeneous mini-batches while updating the same shared student; and (vi) Open-MOPD \citep{yu2026openmopd}, which we reproduce on our model families and task datasets using its proposed capability-balancing method. Details and variants of our MOPD implementation are deferred to Appendix~\ref{app:mopd-mixing}. All OPD comparisons in the main tables use the reverse-KL objective and the same total number of task prompts. We report the task-specific percentage score produced by the same evaluation pipeline for every method and their unweighted average across the three tasks. \method{} uses a Code$\rightarrow$Reason$\rightarrow$Math schedule with four cycles and a batch size of 16. The reported \method{} results are averaged over three independent runs with different random seeds.



\section{Main Results}
\label{sec:results}
\begin{table}[t]
\centering
\vspace{-2.0\baselineskip}
\small
\renewcommand{\arraystretch}{1.05}
\caption{Results across the Qwen2.5-7B and Llama-3.1-8B families. Each group reports task-specific scores and their unweighted average. Bold values mark the best within each model family and column.}
\label{tab:main-results}
\begin{tabular*}{\textwidth}{@{\extracolsep{\fill}}lrrrrrrrr@{}}
\toprule
& \multicolumn{4}{c}{Qwen2.5-7B} & \multicolumn{4}{c}{Llama-3.1-8B} \\
\cmidrule(lr){2-5}\cmidrule(l){6-9}
Model / method & Math & Reason & Code & Avg. & Math & Reason & Code & Avg. \\
\midrule
Student Model & 51.78 & 50.20 & 49.67 & 50.55 & 11.33 & 58.56 & 27.33 & 32.41 \\
Math Teacher & 66.89 & 56.84 & 49.67 & 57.80 & 18.22 & 57.08 & 27.33 & 34.21 \\
Reason Teacher & 58.89 & 70.35 & 50.00 & 59.75 & 8.89 & 68.55 & 24.33 & 33.92 \\
Code Teacher & 64.22 & 60.69 & 58.00 & 60.97 & 11.33 & 56.43 & \best{42.00} & 36.59 \\
\midrule
Parameter Merge & 61.56 & 60.44 & 53.67 & 58.56 & 12.22 & 63.64 & 30.00 & 35.29 \\
MOPD & 65.56 & 71.74 & 56.00 & 64.43 & 18.22 & 68.63 & 30.00 & 38.95 \\
BB-MOPD & 65.33 & 70.68 & 55.67 & 63.89 & 15.33 & 64.21 & 29.33 & 36.29 \\
Open-MOPD & 66.44 & 70.52 & 53.67 & 63.54 & 14.89 & 65.52 & 25.00 & 35.14 \\
\rowcolor{gray!18}
\method{} & \best{67.78} & \best{73.46} & \best{59.67} & \best{66.97} & \best{21.33} & \best{69.45} & 32.33 & \best{41.04} \\
\bottomrule
\end{tabular*}
\vspace{-1.5\baselineskip}
\end{table}

\subsection{Qwen2.5-7B}
Table~\ref{tab:main-results} gives the primary Qwen2.5-7B result. \method{} achieves the best score in every task column and raises the average score across the three tasks to 66.97, outperforming MOPD by 2.54 points and BB-MOPD by 3.08 points. Relative to MOPD, it gains 2.22 points on Math, 1.72 on Reason, and 3.67 on Code. These across-the-board gains demonstrate stronger integration of all three capabilities in a single student.

\subsection{Llama-3.1-8B}
The Llama-3.1-8B columns in Table~\ref{tab:main-results} establish transfer across model families. \method{} achieves the best integrated-model average of 41.04, outperforming MOPD by 2.09 points and BB-MOPD by 4.75 points. Relative to MOPD, it improves Math by 3.11 points, Reason by 0.82 points, and Code by 2.33 points. Improvements on every task confirm that the ordering and projection rule generalize beyond the Qwen configuration and strengthen balanced capability integration across model families.

\subsection{Projection ablation}
\begin{wraptable}{r}{0.60\textwidth}
\centering
\vspace{-2.0\baselineskip} 
\caption{Projection ablation on Qwen2.5-7B under the same order and cycle schedule. The variants isolate the contributions of gradient projection and optimizer-update projection.}
\label{tab:projection-ablation}
\small
\setlength{\tabcolsep}{2.7pt}
\renewcommand{\arraystretch}{1.05}
\begin{tabular}{@{}lrrrr@{}}
\toprule
Variant & Math & Reason & Code & Avg. \\
\midrule
No projection & 67.33 & 68.55 & 55.67 & 63.85 \\
\rowcolor{gray!9}
+ Gradient projection & 67.11 & \best{73.87} & 57.67 & 66.22 \\
\rowcolor{gray!18}
+ Optimizer-update projection & \best{67.78} & 73.46 & \best{59.67} & \best{66.97} \\
\bottomrule
\end{tabular}
\vspace{-0.7\baselineskip} 
\end{wraptable}
We conduct a controlled ablation to separate the effects of the two projection stages from the shared training schedule. All variants use the same Code$\rightarrow$Reason$\rightarrow$Math order, four cycles, task data, and optimization budget. As shown in Table~\ref{tab:projection-ablation}, gradient projection raises the average from 63.85 to 66.22, with particularly clear gains on Reason and Code, showing that removing components aligned with protected task directions substantially mitigates cross-task interference. Projecting the preconditioned optimizer update further improves the average to 66.97 and yields the strongest Math and Code results. Overall, the complete \method{} pipeline outperforms the matched no-projection variant by 3.12 points, supporting the complementary roles of gradient-space protection and optimizer-update correction.

\section{Analysis}
\label{sec:analysis}
We derive lightweight geometric diagnostics for selecting the two principal scheduling choices of \method{}: task order and the number of cycles. Exhaustive search over either choice becomes increasingly costly as the number of tasks or the training scale grows. We therefore connect aggregate evaluation performance to lightweight geometric diagnostics and use them to provide practical, low-cost guidance for selecting an order and cycle count before a full MOPD run.

\subsection{Selecting task order by average undirected conflict}
\begin{wraptable}{r}{0.58\textwidth}
\vspace{-1.0\baselineskip}
\caption{One-cycle performance of all six task orders. C, R, and M denote Code, Reason, and Math, respectively.}
\label{tab:order}
\centering
\small
\setlength{\tabcolsep}{3.4pt}
\begin{tabular}{@{}lrrrr@{}}
\toprule
Order & Math & Reason & Code & Avg.\\
\midrule
C$\rightarrow$M$\rightarrow$R & 65.78 & 74.04 & 55.67 & 65.16\\
M$\rightarrow$R$\rightarrow$C & 69.11 & 70.68 & 55.33 & 65.04\\
C$\rightarrow$R$\rightarrow$M & 70.22 & 73.87 & 53.33 & \best{65.81}\\
M$\rightarrow$C$\rightarrow$R & 66.22 & 69.70 & 54.67 & 63.53\\
R$\rightarrow$C$\rightarrow$M & 65.11 & 72.32 & 56.67 & 64.70\\
R$\rightarrow$M$\rightarrow$C & 64.22 & 72.97 & 51.33 & 62.84\\
\bottomrule
\end{tabular}
\vspace{-1.0\baselineskip}
\end{wraptable}
The three tasks can be arranged in six possible orders. We evaluate all six under the same one-cycle training and data budget, with the results reported in Table~\ref{tab:order}. Their averages span 2.97 points, demonstrating that task order materially affects multi-task integration. Code$\rightarrow$Reason$\rightarrow$Math (C$\rightarrow$R$\rightarrow$M) achieves the highest average of 65.81. To explain why this order is preferable and to avoid exhaustive enumeration in larger task sets, we introduce an average undirected conflict score.

For a memory $M_a$ extracted from task $a$ and a gradient $G_b$ measured on task $b$, we first define the directional conflict
\begin{equation}
r_{a\rightarrow b}=\frac{\|G_bM_aM_a^\top\|_F}{\|G_b\|_F}.
\label{eq:directional-conflict}
\end{equation}
Because this quantity is asymmetric, we symmetrize each task pair and average its conflicts with the remaining tasks:
\begin{equation}
c(a,b)=\tfrac{1}{2}\left(r_{a\rightarrow b}+r_{b\rightarrow a}\right),
\qquad
c(a)=\frac{1}{|\mathcal{T}|-1}\sum_{b\ne a}c(a,b).
\label{eq:task-conflict}
\end{equation}

\begin{wrapfigure}{r}{0.49\textwidth}
\centering
\vspace{-1.7\baselineskip}
\includegraphics[width=\linewidth]{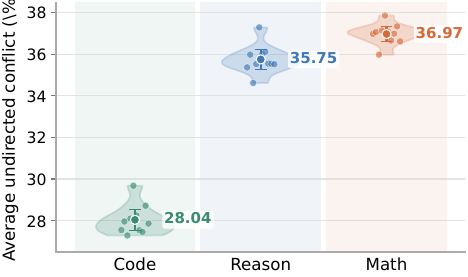}
\caption{Ten-fold probe estimates of task-level average undirected conflict using 60 examples per fold. Shaded regions distinguish the three tasks. Circles show individual folds, diamonds and adjacent values denote fold means, and error bars indicate 95\% confidence intervals. All folds recover Code $<$ Reason $<$ Math.}
\label{fig:order-probe}
\vspace{-1.8\baselineskip}
\end{wrapfigure}

The resulting task-level average undirected conflict scores are 22.34\% for Code, 28.50\% for Reason, and 29.38\% for Math. Sorting them from low to high gives Code$\rightarrow$Reason$\rightarrow$Math, exactly matching the strongest order in the exhaustive ablation.

This ordering is consistent with the mechanism of projected protection. Once a task has been written to memory, subsequent gradients must discard components aligned with its protected subspace. Placing lower-conflict tasks earlier keeps the accumulated memory less broadly conflicting with later optimization, thereby limiting unnecessary removal while still suppressing components that overlap with protected directions. 

We evaluate whether the conflict ranking can be recovered from a lightweight probe rather than full training data. Using 60 examples per fold, Figure~\ref{fig:order-probe} shows the fold-level estimates together with their means and 95\% confidence intervals. All ten folds preserve Code $<$ Reason $<$ Math, with mean estimates of 28.04\%, 35.75\%, and 36.97\%, respectively. The probe therefore recovers the task-level ranking from a compact sample and selects the best observed order before full training, avoiding exhaustive permutation evaluation. The fold-level values and full directional measurements are reported in Appendix~\ref{app:directional}.

\subsection{Selecting the number of cycles by subspace consistency}
\begin{figure}[htbp]
\centering
\includegraphics[width=0.88\columnwidth]{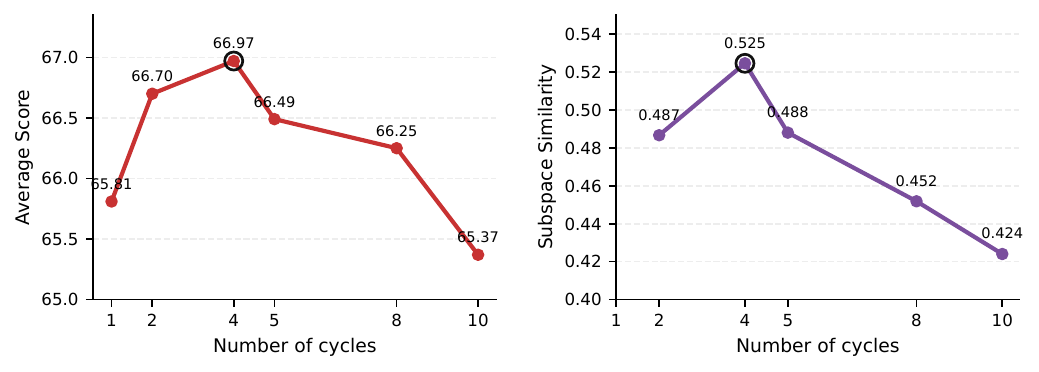}
\caption{Cycle-count analysis under a fixed per-task data budget. (a) Average score across Math, Reason, and Code for the tested schedules, with four cycles achieving the best result. (b) Mean same-task subspace similarity across cycle revisits, computed using Equation~\ref{eq:overlap}. Four cycles also produce the most consistent subspace estimates. Similarity is undefined for the one-cycle schedule because no cross-cycle comparison is available.}
\label{fig:cycle}
\end{figure}

Under a fixed per-task data budget, training can be partitioned into different numbers of cycles. We evaluate 1, 2, 4, 5, 8, and 10 cycles while keeping the task order and total number of prompts unchanged. Figure~\ref{fig:cycle}(a) shows a non-monotonic trend: the average score across the three tasks increases from 65.81 with one cycle to a maximum of 66.97 with four cycles, then decreases to 65.37 with ten cycles. We next examine why the intermediate schedule performs best.

We use same-task cross-cycle subspace similarity as a diagnostic of estimation consistency. Specifically, Equation~\ref{eq:overlap} compares the top-$K$ update subspaces estimated for the same task across consecutive cycle visits, and we average the scores across revisits and tasks. A larger value indicates that repeated estimates recover more consistent update directions and therefore provides a proxy for the reliability of the subspace memory. Figure~\ref{fig:cycle}(b) follows the same pattern as the average evaluation score: similarity is 0.487 for two cycles, peaks at 0.525 for four cycles, and then decreases to 0.488, 0.452, and 0.424 for 5, 8, and 10 cycles, respectively.

The sweep reveals a two-sided scheduling trade-off. At high cycle counts, shorter task blocks provide less data for each cumulative displacement to stabilize, and more frequent memory reconstruction reduces agreement between successive estimates. At low cycle counts, longer uninterrupted blocks increase trajectory drift between visits to the same task. Four cycles balance reliable subspace estimation with timely task revisitation, producing both the highest consistency and the highest average evaluation score.

The coincidence of the performance and consistency peaks at four cycles supports the intended mechanism of projected protection: \method{} performs best when its protected subspaces are estimated most consistently. Detailed per-task similarities are provided in Appendix~\ref{app:revisit-consistency}.

\section{Conclusion}
\label{sec:conclusion}
MOPD interference follows a compact geometric structure: task updates rapidly concentrate in distinct low-dimensional subspaces, and their overlapping components provide a direct target for interference control. \method{} exploits this structure by constructing memories from realized task displacements and projecting both gradients and adaptive-optimizer updates away from protected directions. Combined with conflict-guided task ordering and cycle selection, this mechanism improves every evaluated capability over MOPD and raises the average score across the three tasks by 2.54 points on Qwen2.5-7B and 2.09 points on Llama-3.1-8B. These results establish parameter-update subspace protection as an effective and transferable strategy for integrating specialized teachers into a single balanced model.

\section*{AI Usage Disclosure}
Generative AI tools were used during manuscript preparation to polish the English writing and to assist with the retrieval and discovery of relevant literature. The authors reviewed and revised all AI-assisted outputs, retained full control over the scientific content and citation selection, and take responsibility for the final manuscript.

\bibliography{iclr2027_conference}
\bibliographystyle{iclr2027_conference}

\appendix
\section{Additional Results and Diagnostics}
\subsection{Conflict estimates and directional measurements}
\label{app:directional}

Table~\ref{tab:fold-conflict} reports the task-level average undirected conflict estimated from each of the ten 60-example folds used in Figure~\ref{fig:order-probe}. Every fold recovers the same ordering, Code $<$ Reason $<$ Math, demonstrating that the ranking is stable across compact probe subsets.

\begin{table}[H]
\centering
\caption{Task-level average undirected conflict estimated from each 60-example fold. The final column gives the ascending task order within each fold.}
\label{tab:fold-conflict}
\setlength{\tabcolsep}{5pt}
\begin{tabular}{@{}crrrl@{}}
\toprule
Fold & Code & Reason & Math & Ascending order\\
\midrule
1  & 27.55\% & 35.37\% & 36.98\% & Code $<$ Reason $<$ Math\\
2  & 27.96\% & 35.98\% & 37.07\% & Code $<$ Reason $<$ Math\\
3  & 27.29\% & 34.62\% & 35.98\% & Code $<$ Reason $<$ Math\\
4  & 28.10\% & 35.53\% & 37.14\% & Code $<$ Reason $<$ Math\\
5  & 29.68\% & 37.29\% & 37.86\% & Code $<$ Reason $<$ Math\\
6  & 28.25\% & 36.00\% & 37.09\% & Code $<$ Reason $<$ Math\\
7  & 27.54\% & 36.12\% & 36.66\% & Code $<$ Reason $<$ Math\\
8  & 27.46\% & 35.55\% & 36.99\% & Code $<$ Reason $<$ Math\\
9  & 28.72\% & 35.54\% & 37.34\% & Code $<$ Reason $<$ Math\\
10 & 27.86\% & 35.52\% & 36.62\% & Code $<$ Reason $<$ Math\\
\midrule
Mean & 28.04\% & 35.75\% & 36.97\% & Code $<$ Reason $<$ Math\\
\bottomrule
\end{tabular}
\end{table}

Table~\ref{tab:directional-conflict} reports the asymmetric measurements used to construct the undirected task scores. Here $a\rightarrow b$ means that a memory is first constructed from task $a$ and the removed-gradient ratio is then measured while optimizing task $b$. The two directions differ for every pair, confirming that task conflict cannot be represented by a single symmetric quantity before aggregation.

\begin{table}[H]
\centering
\caption{Directional conflict measured by the removed-gradient ratio.}
\label{tab:directional-conflict}
\begin{tabular}{@{}lc@{}}
\toprule
Direction & Removed gradient\\
\midrule
Math$\rightarrow$Reason & 36.86\%\\
Reason$\rightarrow$Math & 34.22\%\\
Code$\rightarrow$Math & 26.56\%\\
Code$\rightarrow$Reason & 25.06\%\\
Math$\rightarrow$Code & 19.87\%\\
Reason$\rightarrow$Code & 17.85\%\\
\bottomrule
\end{tabular}
\end{table}

\subsection{Per-task revisit consistency}
\label{app:revisit-consistency}

\begin{table}[H]
\centering
\caption{Same-task subspace similarity between cycle revisits.}
\label{tab:per-task-revisit}
\setlength{\tabcolsep}{6pt}
\begin{tabular}{@{}rrrrr@{}}
\toprule
Cycles & Code & Reason & Math & Mean\\
\midrule
2 & 0.486 & 0.603 & 0.371 & 0.487\\
4 & \best{0.535} & \best{0.622} & \best{0.417} & \best{0.525}\\
5 & 0.472 & 0.610 & 0.382 & 0.488\\
8 & 0.433 & 0.584 & 0.338 & 0.452\\
10 & 0.411 & 0.546 & 0.315 & 0.424\\
\bottomrule
\end{tabular}
\end{table}

The four-cycle schedule achieves the highest revisit similarity for every task as well as the highest mean, providing task-wise support for the consistency criterion used to select the cycle count.

\subsection{MOPD mixing variants}
\label{app:mopd-mixing}

\begin{table}[H]
\centering
\caption{Task-mixing variants of MOPD on Qwen2.5-7B.}
\label{tab:mixed-variants}
\setlength{\tabcolsep}{5pt}
\begin{tabular}{@{}lrrrr@{}}
\toprule
Variant & Math & Reason & Code & Avg.\\
\midrule
WB random & 66.67 & 71.91 & 55.00 & 64.52\\
WB ordered & 68.78 & 72.71 & 53.00 & 64.83\\
BB random & 65.33 & 70.68 & 55.67 & 63.89\\
BB ordered & 65.56 & 69.86 & 56.67 & 64.03\\
\bottomrule
\end{tabular}
\end{table}

We additionally compare several task-mixing implementations. Within-batch (WB) mixing places examples from different tasks in the same mini-batch, whereas between-batch (BB) mixing interleaves task-homogeneous mini-batches. ``Random'' and ``ordered'' indicate how tasks or samples are arranged within the corresponding scheme. The strongest mixing variant reaches 64.83, while the complete \method{} configuration reaches 66.97, preserving a 2.14-point advantage over task mixing alone.

\end{document}